\documentclass{article}

\usepackage{arxiv}

\usepackage[utf8]{inputenc} 
\usepackage[T1]{fontenc}    
\usepackage{hyperref}       
\usepackage{url}            
\usepackage{booktabs}       
\usepackage{amsmath}
\usepackage{amssymb}
\usepackage{amsfonts}       
\usepackage{nicefrac}       
\usepackage{microtype}      
\usepackage{cleveref}       
\usepackage{lipsum}         
\usepackage{graphicx}
\usepackage{natbib}
\usepackage{doi}

\usepackage{tikz}
\usepackage{comment}
\usepackage{mathtools}
\usepackage{color, colortbl}
\usepackage{xcolor}
\usepackage{mathtools}
\usepackage{placeins}
\usepackage{multirow}
\usepackage{multicol}
\usepackage{tabularx}
\usepackage{etoc}
\DeclareMathOperator*{\argmin}{argmin}

\title{A Human-centered Explainable AI Framework for Burn Depth Characterization}

\author{ Maxwell J. Jacobson \\
	Purdue University\\
	\texttt{jacobs57@purdue.edu} \\
	\And
	Daniela Chanci Arrubla \\
    Emory University \\
    \texttt{daniela.chanci.arrubla@emory.edu} \\
    \And
    Maria Romeo Tricas \\
    Purdue University \\
    \texttt{mromeotr@purdue.edu} \\
    \And
    Gayle Gordillo \\
    Indiana University \\
    \texttt{gmgordil@iu.edu} \\
    \And
    Yexiang Xue \\
    Purdue University \\
    \texttt{yexiang@purdue.edu} \\
    \And
    Chandan Sen \\
    Indiana University \\
    \texttt{cksen@iu.edu} \\
    \And
    Juan Wachs \\
    Purdue University \\
    \texttt{jpwachs@purdue.edu} \\
}

\renewcommand{\shorttitle}{Human-centered XAI for Burn Depth Characterization}

\begin{document}
\maketitle

\begin{abstract}
Approximately 1.25 million people in the United States are treated each year for burn injuries. Precise burn injury classification is an important aspect of the medical AI field. In this work, we propose an explainable human-in-the-loop framework for improving burn ultrasound classification models. Our framework leverages an explanation system based on the LIME classification explainer to corroborate and integrate a burn expert's knowledge --- suggesting new features and ensuring the validity of the model. Using this framework, we discover that B-mode ultrasound classifiers can be enhanced by supplying textural features. More specifically, we confirm that texture features based on the Gray Level Co-occurance Matrix (GLCM) of ultrasound frames can increase the accuracy of transfer learned burn depth classifiers. We test our hypothesis on real data from porcine subjects. We show improvements in the accuracy of burn depth classification --- from ~88\% to ~94\% --- once modified according to our framework. 
\end{abstract}

\section{Introduction}

\begin{figure}[tbh]
\centering
\includegraphics[width=.9\textwidth]{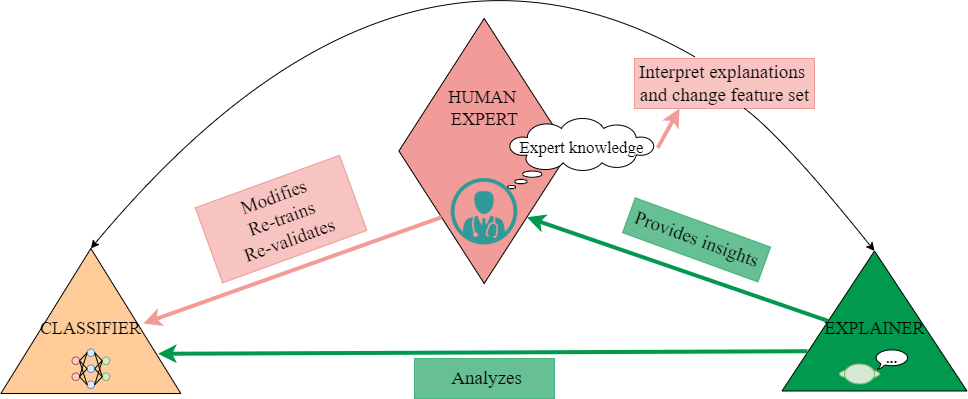}
\caption{Our human-in-the-loop through explainability framework. A human possessing expert knowledge trains a classifier. Our LIME-based explainer  provides insights to the human based on the current model. The human applies expert knowledge to interpret the explanation. The human expert modifies the model appropriately, and the process can be repeated. In our case, we modify the network by adding new features based on expert medical knowledge regarding burns.}
\label{fig:burnhitl}
\end{figure}


Our work focuses on the task of burn depth estimation. Performing this task accurately can be critical for the welfare of burn victims, but it is also a challenge due to the high variability in the visual appearance of burns. Different imaging modalities have been explored to solve this problem and improve the accuracy of the diagnosis \cite{sen_cutaneous_2016} --- e.g. color photographs, ultrasound, infrared thermography, laser speckle imaging, and laser doppler imaging \cite{thatcher_imaging_2016}. Recent studies combine imaging modalities with machine learning and deep learning models. Cirillo et al \cite{cirillo_tensor_2019} compared the performance of VGG16 \cite{vgg}, GoogleNet \cite{googlenet}, ResNet50 \cite{resnet}, and ResNet101, pretrained on ImageNet \cite{imagenet}, to classify their labeled dataset of RGB burn images. Similarly, Chauhan \& Goyal \cite{chauhan_convolution_2020-1} carried out the burn depth classification based on the specific characteristics of the body region in which the injury was located. However, RGB images can negatively influence the accuracy of predictions due to lighting conditions, skin color, or general variability in wound presentation. Besides, RGB images, optical coherence tomography \cite{singla_vivo_2018}, spatial frequency-domain imaging \cite{rowland_burn_2019}, and ultrasound \cite{lee_real-time_2020-1} have been utilized for feature extraction. Harmonic B-mode ultrasound (HUSD) --- a non-invasive sound-based imaging technique --- is used in this work. B-mode ultrasound is based on the transmission of small pulses of ultrasound. Echoes reflected back to the transducer from body tissues that have different acoustic impedances, which can be measured to build a 3D map of tissues. Second harmonic frequency echoes are used in order to reduce the artifacts in the image produced by the reflection of echoes at different frequencies \cite{narouze_atlas_2018}.

Computer vision based techniques lack the human expertise in current medical ML models. Therefore, a human-in-the-loop system built on Explainable Artificial Intelligence is proposed here. Human-in-the-loop (HITL) is an Artificial Intelligence (AI) paradigm that assumes the presence of human experts that can guide the learning or operation of the otherwise-autonomous system. Lundberg et al. \cite{lundberg_hypoxaemia_2018} developed and tested a system to prevent hypoxaemia during surgery by providing anaesthesiologists with interpretable hypoxaemia risks and contributing factors. Later, Sayres et al. \cite{DR_2019} proposed and evaluated a system to assist diabetic retinopathy grading by ophtalmologists using a deep learning model and integrated gradients explanation.

An Explainable Artificial Intelligence (XAI) is an intelligent system which can be explained and understood by a human \cite{gohel_AIX_2021}. For this, we utilize LIME (Local Interpretable Model-agnostic Explanations)  \cite{lime}, a recent method that is able to explain the predictions of any classifier model in an interpretable manner. LIME operates by roughly segmenting an image into feature regions, then assigning saliency scores to each region. Higher scoring zones are more important in arriving at the classification result of the studied model. The algorithm first creates random permutations of the image to be explained. Then, the classification model is run on those samples. Distances between those samples and the original image can be calculated, which are then converted to weights by being mapped between zero and one using a kernel function. Finally, a simple linear model is fitted around the initial prediction to generate explanations. This explanation provided by LIME is the result of the following minimization:
\begin{equation}
\xi (x) = \argmin_{g \in G} \mathcal{L}(f,g,\pi_x) + \Omega(g)
\end{equation}
Let $x$ and $f$ be the image and classifier to be examined and $G$ as the class of interpretable models like decision trees and linear models. The complexity of the model (e.g. depth of a decision tree, number of non-zero weights in a linear model) should be as small as possible to maintain explainability. This complexity can be defined as $\Omega(g)$. Explanations by LIME are found by fitting explainable models --- minimizing the sum of the local faithfulness loss $L$ and the complexity score. Permutated sampling is used to approximate this local faithfulness loss. For this reason, a proximity measure $\pi_x(z)$ calculates the distance between $x$ and another image $z$. The objective is minimizing the fidelity function while maintaining a measure of complexity low enough to be interpretable. This minimization makes no assumptions about $f$ in order to generate a model-agnostic explanation. 

Explainable intelligence is useful when combined with HITL systems because they provide understandable and qualitative information about the relationship between the instance's components and the model's prediction. Therefore, an expert can make an informed decision about whether the model is reliable, and can make the necessary changes if it is not --- eventually reaching a confident result. This is extremely important, above all in the medical field, because of the severe ethical implications that suppose a wrong medical diagnosis. 

By deploying our explainable human-in-the-loop method, we were able to confirm the importance of one family of features which can enhance a convolutional burn prediction classifier --- statistical texture.
From the Gray Level Co-ocurrence Matrix (GLCM) --- a method that represents the second-order statistical information of gray levels between neighboring pixels in an image --- many important sub-features can be extracted \cite{seal_2018}. In this work, the features used are: contrast, homogeneity, angular second moment (ASM), energy, and dissimilarity. More information on GLCM can be found in appendix A.

We find that by discovering weaknesses in our model through explainable AI and utilizing expert knowledge from a human in the loop, burn depth can be classified more accurately from B-mode ultrasound data.

\begin{figure}[tb]
\centering
\includegraphics[width=.9\textwidth]{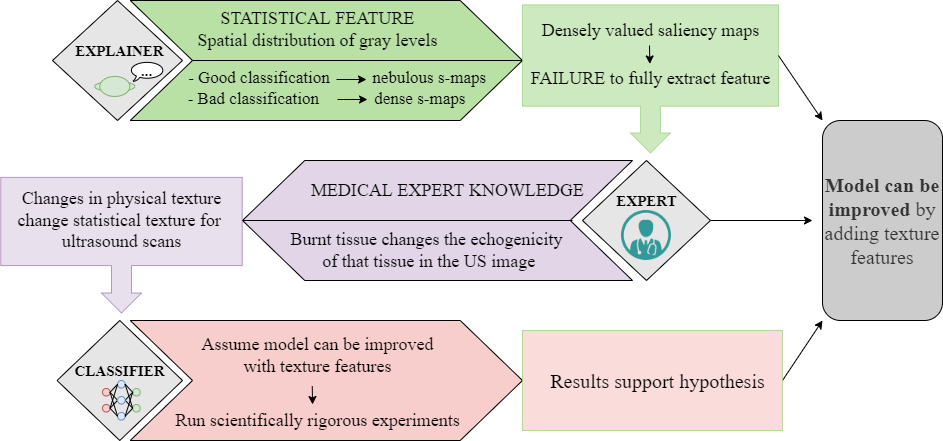}
\caption{When applying our framework to the question of texture in ultrasound burn analysis, we can show support for the hypothesis in three ways: from explanation, from expert knowledge, and from scientific experimentation. This allows for greater understanding and trust in the classifier.}
\label{fig:burn_support}
\end{figure}

\section{Method}
\FloatBarrier
\begin{figure}[tb]
\centering
\includegraphics[width=0.95\textwidth]{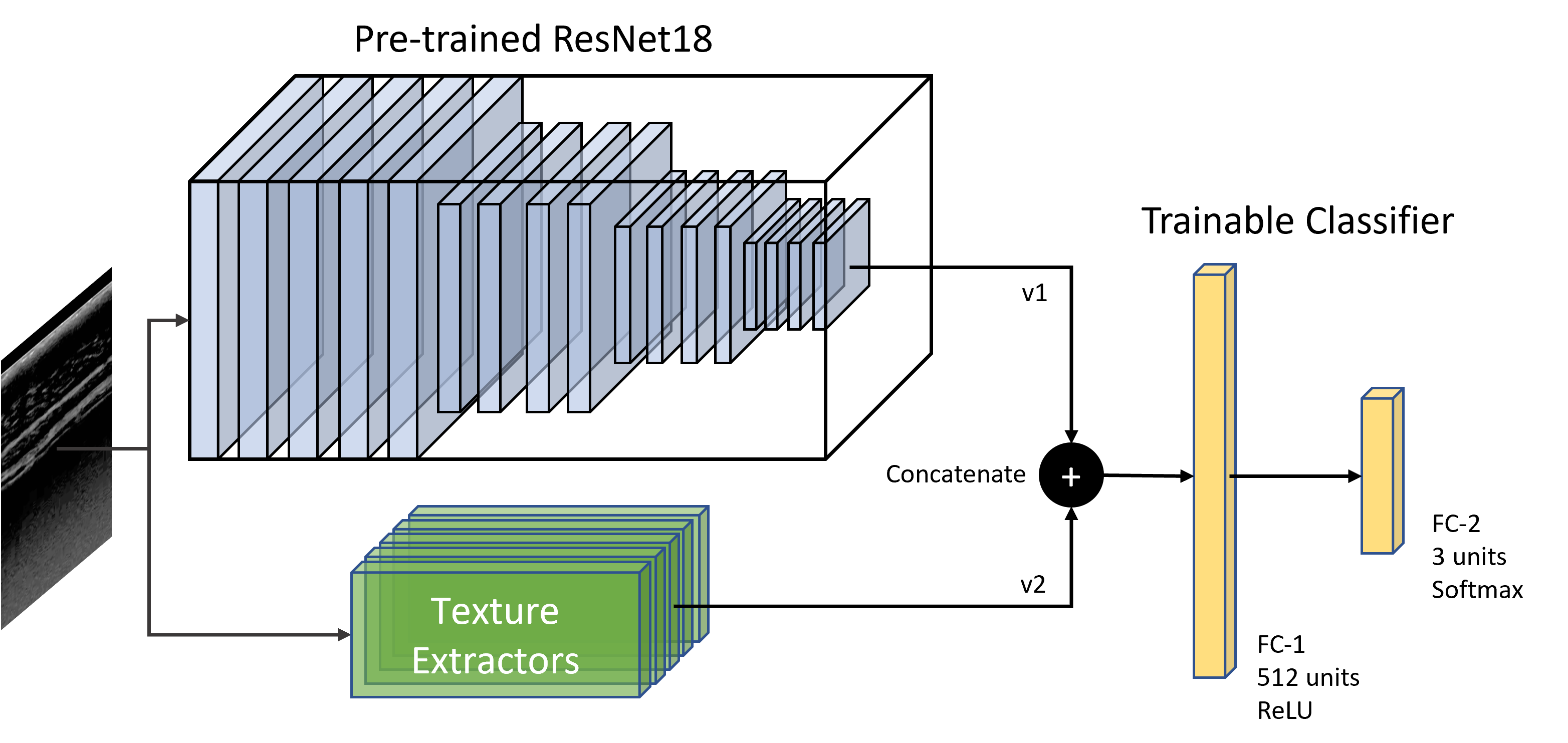}
\caption{Burn classification model. The primary feature vector ($v_1$) is calculated using the pre-trained model. This model is not updated during training. The secondary vector ($v_2$) is calculated from GLCM texture features. Both are concatenated as inputs to a simple 2-layer Softmax classifier. This layer contains 1024 unit ReLU layer before the output layer.}
\label{fig:arch}
\end{figure}

\textbf{Explainable Human-in-the-loop System.} The primary contribution of this paper is our XAI-based human-in-the-loop system (See figure \ref{fig:burnhitl}). The human expert starts the process --- in this case, the expert was a clinician on the team. They train and test an initial classifier model on a burn ultrasound dataset. If this does not yield satisfactory results, the expert activates the explainer system. Using hand-selected or random records from the dataset, the LIME-based explainer gives a series of explanations about the model to the expert. These explanations are in the form of saliency maps --- visually communicating to the expert what parts of the input are being utilized for any given prediction.

For our purposes, LIME conducts its sampling using the quickshift \cite{quickshift} and Felzenszwalb \cite{felzenszwalb} segmentation methods. 10,000 permutations are sampled, and the LIME algorithm is run. From the LIME outputs, we create a heatmap, a visual overlay of the top K features, and a quantitative list of the LIME scores for each feature. Besides LIME, we also produce a pixel-wise saliency map using back-propagation on the input image \cite{salmap}. This approach is not model agnostic, but it provides a second explanation modality to include when the classifier is a neural net.

\begin{figure}[tb]
\centering
\includegraphics[width=0.75\textwidth]{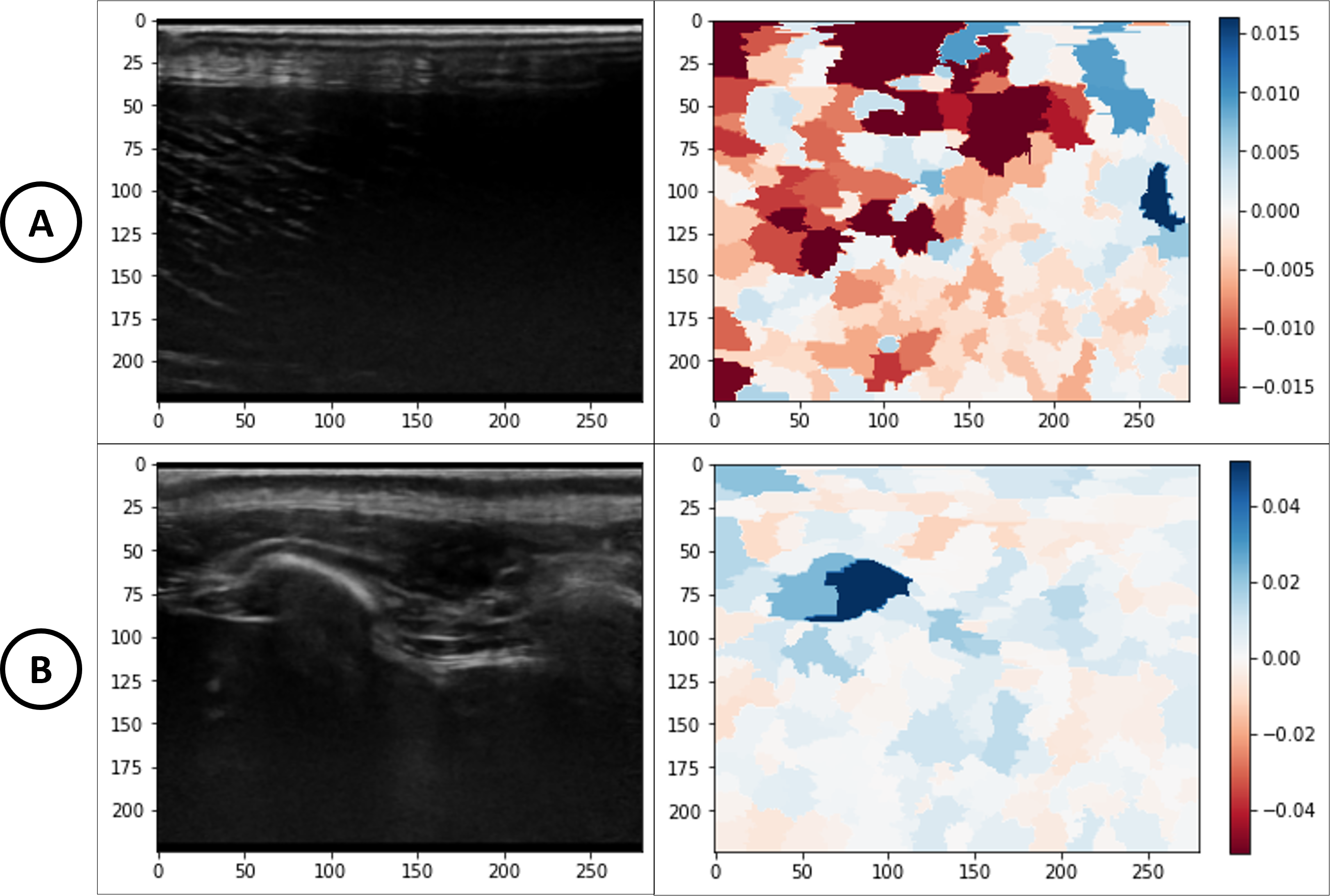}
\caption{Demonstration of explainer output. \textbf{A)} Shows a correct classification. While certain zones of the image are more salient than others, information is utilized from across the image. It is likely that texture information is being naturally extracted. Note for scale that LIME scores range between -0.027 and 0.016. \textbf{B)} Shows a misclassified example. The output is largely determined by one feature of the input. The correct texture features are either not being calculated or are being ignored. Note for scale that LIME scores range between -0.061 and 0.012 --- A nearly 70\% increase.}
\label{fig:ab}
\end{figure}

The expert combines this new information with their prior medical knowledge. In this case, the burn expert knows that severe burns change the echogenicity of healthy skin. Further, the expert knows that physical textural features in ultrasound translate to statistical textural changes in the image \cite{lee_real-time_2020-1}. This can be interpreted and implies that, in addition to key areas like skin layer transitions, a solid classifier should have a nebulous saliency map, extracting features from across the image. In our case, the expert was provided with these explanations and observed dense saliency maps were more apparent on misclassified examples. This view is demonstrated in Figure \ref{fig:ab}. 

After the interaction between the explainer and the expert, the necessary modifications in the feature set of the convolutional neural network can be made. In this work, this is done by adding GLCM texture features --- selected in the hope that they can assuage the "feature bottleneck" and improve accuracy. Finally, the model is re-trained and re-tested to prove that the updated classifier leads to a good prediction. Consequently, this human-in-the-loop system allows the design of a trustworthy, ethical, and robust medical diagnosis system (See figure \ref{fig:burn_support}.

\textbf{Classification Model.} In this work, ResNet18 is used as a pretrained ---on ImageNet dataset--- convolutional neural network with all his layers frozen except from the last fully connected layer, which is replaced by a 1-layer softmax classifier. The hyperparameters set for this model are: 15 epochs, a batch size equal to 8 and a $10^{-5}$ learning rate with no decay. The cross-entropy loss function was utilized alongside the Adam optimizer \cite{adam} for training.
The B-mode ultrasound imaging modality consists of grayscale images with different textures, which are very important in the diagnostic process. Therefore, in addition to the presented architecture, a feature approach is followed in this work (See figure \ref{fig:arch}) . The last layer of the ResNet18 neural network outputs a 30-dimension feature vector from the 512 features that the architecture initially extracts. This vector ($v_1$) constitutes the primary feature vector, which is combined with the hand-crafted Haralick texture features vector ($v_2$). Hence, the Haralick features extracted from the GLCM matrix are concatenated with the CNN-based features, and afterwards, the fused vector constitutes the input to the fully connected softmax classifier.

\section{Data}
An unreleased B-mode HUSD dataset was acquired for each of the three classes included in the burn depth classification task: full-thickness burn, partial-thickness burn, and unburnt skin. Pig models were selected as the \textit{in vivo} wound-healing animal models based on previous studies, which indicate that there is a high similarity between pig skin and human skin. Similar characteristics include the dermal-epidermal ratio, the dermal collagen, the distribution of blood vessels, and an abundant subdermal adipose tissue. In addition, the healing process is also comparable, more specifically, it was demostrated that there is a similarity of 78\% between pigs and humans. This number decreases to 53\% for small-mammal models, and 57\% for in vitro studies \cite{sullivan_pig_2001}. 
B-mode HUSD videos of approximately 15 seconds were obtained. The data corresponding to the skin class was collected first, in the same region where the burn injuries were going to be located. Then, in order to collect the data for the two remaining classes, eight full-thickness burn injuries were generated. The data for the full-thickness class was collected immediately after wounding (day 0). The wounds were then treated, and the healing process started. By day 14 post-burn, the injuries had not healed completely. More specifically, by that time the subcutaneous tissue had been regenerated, while the dermis and the epidermis were still affected. These are the same characteristics that can be seen in second-degree burn injuries, reason why day 14 post-burn data was used for the second-degree class.

\section{Experiments \& Results}

\begin{table}[p]
  \centering
    \begin{tabular}{p{8em}|c c c c c c c c}
    \toprule
    {\textbf{Features}} & \multicolumn{1}{p{3.2em}}{ {\textbf{Mean \newline Acc}}} & \multicolumn{1}{p{3.2em}}{ {\textbf{Acc \newline Std}}} & \multicolumn{1}{p{3.2em}}{ {\textbf{Mean \newline Prec}}} & \multicolumn{1}{p{3.2em}}{ {\textbf{Prec \newline Std}}} & \multicolumn{1}{p{3.2em}}{ {\textbf{Mean \newline Rec}}} & \multicolumn{1}{p{3.2em}}{ {\textbf{Rec \newline Std}}} & \multicolumn{1}{p{3.2em}}{ {\textbf{Mean F1}}} & \multicolumn{1}{p{3.2em}}{ {\textbf{F1 Std}}} \\
     \midrule
     \midrule
      SVM & 91.90\% & 0.016 & 0.852 & 0.045 & \textbf{0.935} & 0.014 & 0.874 & 0.036 \\
     \midrule
      ResNet18 & 87.53\% & 0.014 & 0.872 & 0.014 & 0.848 & 0.015 & 0.853 & 0.015 \\
     \midrule
      ResNet18 +\\ Contrast & \textbf{93.62\%} & 0.014 & \textbf{0.934} & 0.011 & 0.913 & 0.013 & \textbf{0.918} & 0.013 \\
     \midrule
      ResNet18 +\\ Homogeneity & 88.75\% & 0.006 & 0.886 & 0.006 & 0.863 & 0.006 & 0.868 & 0.007 \\
     \midrule
      ResNet18 +\\ ASM & 88.54\% & 0.007 & 0.883 & 0.011 & 0.860 & 0.012 & 0.865 & 0.011 \\
     \midrule
      ResNet18 +\\ Energy & 88.54\% & 0.005 & 0.881 & 0.009 & 0.856 & 0.009 & 0.862 & 0.009 \\
     \midrule
      ResNet18 +\\ Dissimilarity & 89.56\% & 0.010 & 0.890 & 0.009 & 0.868 & 0.010 & 0.874 & 0.010 \\
     \midrule
      ResNet18 +\\ All Features & 92.56\% & \textbf{0.004} & 0.927 & \textbf{0.005} & 0.903 & \textbf{0.004} & 0.909 & \textbf{0.004} \\
    \bottomrule
    \end{tabular}%
  \vspace{2pt}
  \caption{Results on burn depth classification. Each method was trained and tested 6 times. 30 Epochs of training were conducted using a random subset of the training data containing 1,400 frames per class. Pure ResNet18 is compared with the feature-augmented versions, as well as an SVM due to its common use in this domain. All GLCM features show significant improvement over pure ResNet18. Furthermore, only the Contrast feature shows very high improvement. Although, combining all 5 features shows a similar improvement but with lower variability.}
  \label{tab:res}%
\end{table}%

\begin{table}[p]
\centering
\begin{minipage}[t]{.49\textwidth}
  \centering
    \begin{tabular}[t]{l l c c c}
    \toprule
    \multicolumn{5}{c}{\large Tukey Test Results} \\
    \textbf{I} & \textbf{J} & \multicolumn{1}{l}{\textbf{MD}} & \multicolumn{1}{l}{\textbf{P-adj}} & \multicolumn{1}{l}{\textbf{Reject}} \\
    \hline
    \hline
    \multicolumn{1}{c}{\multirow{6}[0]{*}{All}} & ASM   & -0.0402 & 0     & \multicolumn{1}{c}{\checkmark} \\
          \cmidrule{2-5}
          & Contr & 0.0106 & 0.461 &  \\
          \cmidrule{2-5}
          & Dissim & -0.03 & 0     & \multicolumn{1}{c}{\checkmark} \\
          \cmidrule{2-5}
          & Energy & -0.0402 & 0     & \multicolumn{1}{c}{\checkmark} \\
          \cmidrule{2-5}
          & Homog & -0.0381 & 0     & \multicolumn{1}{c}{\checkmark} \\
          \cmidrule{2-5}
          & None  & -0.0503 & 0     & \multicolumn{1}{c}{\checkmark} \\
    \hline
    \multicolumn{1}{c}{\multirow{5}[0]{*}{ASM}} & Contr & 0.0508 & 0     & \multicolumn{1}{c}{\checkmark} \\
          \cmidrule{2-5}
          & Dissim & 0.0102 & 0.501 &  \\
          \cmidrule{2-5}
          & Energy & 0     & 1     &  \\
          \cmidrule{2-5}
          & Homog & 0.0021 & 1     &  \\
          \cmidrule{2-5}
          & None  & -0.0101 & 0.514 &  \\
    \bottomrule
    \end{tabular}
    \end{minipage}
    \begin{minipage}[t]{.49\textwidth}
    \centering
    \begin{tabular}[t]{l l c c c}
    \toprule
    \multicolumn{5}{c}{\large Tukey Test Results Continued} \\
    \textbf{I} & \textbf{J} & \multicolumn{1}{l}{\textbf{MD}} & \multicolumn{1}{l}{\textbf{P-adj}} & \multicolumn{1}{l}{\textbf{Reject}} \\
    \hline
    \hline
    \multicolumn{1}{c}{\multirow{4}[0]{*}{Contr}} & Dissim & -0.0406 & 0     & \multicolumn{1}{c}{\checkmark} \\
          \cmidrule{2-5}
          & Energy & -0.0508 & 0     & \multicolumn{1}{c}{\checkmark} \\
          \cmidrule{2-5}
          & Homog & -0.0487 & 0     & \multicolumn{1}{c}{\checkmark} \\
          \cmidrule{2-5}
          & None  & -0.0609 & 0     & \multicolumn{1}{c}{\checkmark} \\
    \hline
    \multicolumn{1}{c}{\multirow{3}[0]{*}{Dissim}} & Energy & -0.0102 & 0.501 &  \\
          \cmidrule{2-5}
          & Homog & -0.0081 & 0.741 &  \\
          \cmidrule{2-5}
          & None  & -0.0203 & 0.01  & \multicolumn{1}{c}{\checkmark} \\
    \hline
    \multicolumn{1}{c}{\multirow{2}[0]{*}{Energy}} & Homog & 0.0021 & 1     &  \\
          \cmidrule{2-5}
          & None  & -0.0101 & 0.514 &  \\
    \hline
    Homog & None  & -0.0122 & 0.295 &  \\
    \bottomrule
    \end{tabular}%
    \end{minipage}
    \vspace{2pt}
  \caption{Tukey test results for accuracy across each experimental feature method. The rejections are the same for all the metrics. Mean difference, adjusted P values, and null hypothesis rejection status are given for each pair. There does not exist a statistically significant difference between all features and contrast, which are the two methods with better results in table 1. All features and Contrast show statistically significant improvement over other feature sets.}
  \label{tab:tuckey}%
\end{table}%

To test our classifier before and after the HITL inspired improvement, we trained it on our ultrasound dataset for 30 epochs. The basic data record was chosen to be frames instead of full videos. The training set was composed of 1,974 full thickness, 1,802 partial thickness, and 1,493 unburnt skin examples. The testing set was composed of 1,309 full thickness, 986 partial thickness, and 141 unburnt skin examples. Another 1072, 805, and 141 respective samples were set aside for validation and hyper-parameter tuning. Each dataset was composed of frames from unique videos.

Preprocessing consisted of a normalization step to translate image tensors to the [0, 1] domain. A center crop was also performed to isolate the true ultrasound data from scan metadata, which could otherwise contaminate the results. This crop brought all images to the size 800x1,000. Resizing was also done, compressing the image to the size 224x224 --- this being the standard input size to ResNet18.

As this dataset was unbalanced, average F1 score and score standard deviation was collected alongside accuracy. All statistics were averaged over 6 training runs with different random seeds for initialization. To show that no single feature is sufficient to gain the best results, we train models with no extra features (baseline), one extra feature (for each GLCM method), and all features combined. We also train a Support Vector Machine SVM for comparison, as this model is common in the burn analysis domain \cite{HUANG20211691,lee_real-time_2020-1,yadav_feature_2019}. Results can be seen in Table \ref{tab:res} and statistical analysis can be seen in Table \ref{tab:tuckey}. In summary, our results show statistically significant improvements when adding some or all of the GLCM texture features. In particular, contrast and dissimilarity are important additions to our ResNet features.


\FloatBarrier
\section{Conclusions}
This work has introduced a method for improving the accuracy of ultrasound burn classifiers using explainable AI and a human-in-the-loop framework. Our real-data experiments show that human experts can leverage their knowledge and the insights of explainable AI to uncover new "feature bottlenecks". Further, the results of these experiments show that by modifying the network according to these insights, significant improvements can be gained. In our future work, we will be adapting this process to different ultrasound modalities and increasing automation and explainability within the framework.

\FloatBarrier
\section{Acknowledgments}
This work was supported by the Office of the Assistant Secretary of Defense for Health Affairs under Award No. 6W81XWH-21-2-0030, by the National Institutes of Health under award 5R21LM013711-02, and by the National Science Foundation under Grant NSF \#2140612. 
Opinions, interpretations, conclusions and recommendations are those of the authors and are not necessarily endorsed by the Department of Defense or by the National Science Foundation.

\bibliographystyle{plainnat}
\bibliography{references}

\end{document}